\documentclass[12pt]{article}
\usepackage[a4paper,margin=2.5cm]{geometry}
\usepackage{amsmath,amssymb}
\usepackage{graphicx}
\usepackage{booktabs}
\usepackage[super]{cite}
\usepackage[hidelinks]{hyperref}
\usepackage{parskip}

\begin{document}

\begin{center}
{\LARGE\bfseries What the model knows but does not say: monitoring medical LLMs through verbalized thoughts\par}
\vspace{1.5em}
{\large Frederik Hauke\textsuperscript{1,*}, Jan Clusmann\textsuperscript{2}, Hanna Kreutzer\textsuperscript{1}, Christiane Kuhl\textsuperscript{1}, Tianyu Han\textsuperscript{5}, Felix Barajas Ordonez\textsuperscript{1}, Sven Nebelung\textsuperscript{1}, Jakob Nikolas Kather\textsuperscript{2,3,4} \& Daniel Truhn\textsuperscript{1}\par}
\end{center}

\vspace{1em}
{\small
\textsuperscript{1}Department of Diagnostic and Interventional Radiology, University Hospital RWTH Aachen, Aachen, Germany.
\textsuperscript{2}Else Kr\"oner Fresenius Center for Digital Health, TU Dresden, Dresden, Germany.
\textsuperscript{3}Department of Medicine I, Faculty of Medicine and University Hospital Carl Gustav Carus, TUD Dresden University of Technology, Dresden, Germany.
\textsuperscript{4}Department of Medical Oncology, National Center for Tumor Diseases (NCT), Heidelberg University Hospital, Heidelberg, Germany.
\textsuperscript{5}University of Pennsylvania, Philadelphia, PA, USA.
\textsuperscript{*}Corresponding author: Frederik Hauke, Department of Diagnostic and Interventional Radiology, University Hospital RWTH Aachen, Pauwelsstra\ss e 30, 52074 Aachen, Germany; \href{mailto:fhauke@ukaachen.de}{fhauke@ukaachen.de}.\par}

\vspace{1em}
\begin{abstract}
Efforts to keep clinical language models safe have concentrated on what enters and leaves the model, but a manipulated model can return an answer that looks entirely ordinary. Here we show that the model's own internal state, rendered in plain language, exposes manipulations the output conceals. Using a natural-language autoencoder, we translate the activations of a multimodal LLM (Gemma-3-27B) into text and let an input-blind judge read this verbalized internal state (the thoughts channel) and the model's answer separately. Across two clinical attack objectives (omitting a pathology, covertly recommending a drug), two input modalities, and two delivery mechanisms, the thoughts channel caught the injection significantly more often than the answer in all eight injected conditions ($p<0.05$), while both channels held near-zero false-positive rates on clean inputs; the omission result carried over to Llama-3.3-70B. The thoughts channel also registered unsafe clinical requests as harmful while the model was still reading them, before it decided whether to refuse. None of this requires retraining: an off-the-shelf verbalizer and a general-purpose judge are enough.
\end{abstract}

\section{Introduction}

Large language models are increasingly integrated into healthcare applications, where they support patient education, clinical decision support, and preliminary diagnosis \cite{lee_2025_oi_251340_1765554485.53247}, treatment recommendations \cite{s41467_025_64062_1}, and image interpretation and documentation \cite{s41467_024_55631_x}; multi-agent configurations, in which each LLM takes on a specialist role, are being explored for more complex medical scenarios \cite{MedSentry_2025}. Yet their clinical safety profiles remain poorly characterized, and strong results on knowledge benchmarks cannot stand in for an explicit measurement of safety \cite{NOHARM_ClinicalSafety_2025}. Medical LLMs readily comply with harmful requests, and do so more often than their safety-aligned general-purpose counterparts \cite{Han_MedSafetyBench_NeurIPS2024}. An evaluation of 31~LLMs on 100 real primary-care-to-specialist consultation cases found potential for severe harm in up to 22\% of cases, and safety performance was only moderately correlated with existing AI and medical-knowledge benchmarks ($r = 0.61$--$0.64$) \cite{NOHARM_ClinicalSafety_2025}. Because the models are generally accurate and their recommendations read as plausible, clinicians tend to default to accepting them, so such errors are prone to go unnoticed through automation bias \cite{NOHARM_ClinicalSafety_2025}. A separate vulnerability stems from the models themselves: frontier LLMs comply with up to 100\% of illogical medical requests, prioritizing helpfulness over logical consistency and generating false clinical information even when they know the request is flawed \cite{s41746_025_02008_z}.

Model limitations are not the only source of error: model behaviour can also be manipulated deliberately. Prompt injection attacks exploit the fact that LLMs operate directly on text and have no formal way to tell instructions from data \cite{NeurIPS_2024_agentdojo_a_dynamic_environment_to_evaluate_prompt_injection_attacks_and_defenses_for_llm_agents_Paper_Datasets_and_Benchmarks_Track,Greshake_IndirectPI_2023}, so an attacker who controls part of the input can override the intended behaviour of the system \cite{PerezRibeiro_PromptInjection_2022,Greshake_IndirectPI_2023,Liu_FormalizingPI_USENIX2024}. Multimodal models have widened this attack surface further \cite{ImageBasedPromptInjection_2025}. Instructions hidden in medical images at low contrast or in tiny font, not obvious to human observers, can make vision-language models produce harmful output \cite{s41467_024_55631_x}, systematically designed image-based injections succeed in up to 64\% of attempts under stealth constraints \cite{ImageBasedPromptInjection_2025}, and in multi-agent pipelines a single injected prompt can replicate from agent to agent, enabling coordinated data theft or system-wide disruption \cite{PromptInfection_MultiAgent_2025}.

In clinical settings, such attacks carry direct consequences for patient safety \cite{lee_2025_oi_251340_1765554485.53247,s41467_025_64062_1,MPIB_MedicalPromptInjection_2025,AdversarialHallucinationMedical_2025}. In a controlled evaluation of commercial LLMs giving medical advice, prompt injections succeeded in 94.4\% of trials and in 91.7\% of dialogues with extreme harm potential, such as recommending thalidomide during pregnancy, and even flagship models with enhanced safety mechanisms remained susceptible in 80--100\% of cases \cite{lee_2025_oi_251340_1765554485.53247}. This vulnerability is not confined to a particular model class: open-source and proprietary models alike can be compromised through prompt injection and fine-tuning attacks across prevention, diagnosis, and treatment tasks \cite{s41467_025_64062_1}. At the same time, results from the Medical Prompt Injection Benchmark (MPIB) show that attack-success rate and clinical harm can diverge substantially, which argues for measuring patient-level risk rather than instruction compliance alone \cite{MPIB_MedicalPromptInjection_2025}. The problem extends beyond explicit instructions: when fabricated clinical details are planted in a prompt, models elaborate on the false information in 50--82\% of cases \cite{AdversarialHallucinationMedical_2025}.

Most existing defences act on the input, on the surrounding system architecture, or on the model itself, which is then hardened through adversarial training or preference optimization \cite{document}. Input-level filtering detects 60--80\% of injection attempts, and advanced architectural protections reach up to 95\% protection against known attack patterns, yet substantial gaps remain against attack vectors that have not been seen before \cite{document}. Training-free defences that repurpose attack techniques against the injected instruction can push attack success close to zero, but only in certain scenarios \cite{DefensePromptInjection_ACL2025}. Different as they are, all of these defences work from the outside. None of them looks at what is happening inside the model while it runs, yet that is exactly where the signal is. \\
A series of recent studies shows that much of what is needed to catch unsafe behaviour can be read directly from an LLM's hidden states \cite{JailbreakLeavesTrace_2025,HiddenDetect_2025,LLMSafetyFromWithin_2025,PIShield_2025}. Because instruction-tuned models represent data containing injected instructions differently from benign data, a linear classifier on residual-stream vectors detects prompt injection with average false-positive and false-negative rates of 0.5\% and 1.4\%, without fine-tuning the model or generating a response \cite{PIShield_2025}. Jailbreak and benign prompts likewise leave consistent latent-space patterns across several open-source model families \cite{JailbreakLeavesTrace_2025}, vision-language models show distinct activation patterns for unsafe prompts that can be used for detection without extensive fine-tuning \cite{HiddenDetect_2025}, and linear probing across layers isolates safety neurons that, once combined, outperform dedicated guard models with 250 times fewer trainable parameters \cite{LLMSafetyFromWithin_2025}. Sparse autoencoders and representation engineering widen what can be extracted from these activations, decomposing them into interpretable features and steerable directions for high-level properties such as honesty and harmlessness \cite{Bricken_SAE_Monosemantic_2023,Zou_RepresentationEngineering_2023,RepEngSurvey_2025}. Yet each of these detectors is trained for one property on one model, and returns a score rather than something a person can read.

Our work builds on Natural Language Autoencoders (NLAs), a recent approach that removes this limitation. An NLA couples two models, a verbalizer that describes residual-stream activations in plain language and a reconstructor that has to rebuild the activations from that description alone; both are trained jointly by reinforcement learning \cite{NLA_Anthropic_2026}. The training objective is nothing but reconstruction, yet the descriptions that emerge read as plausible accounts of the model's internal state, and they have already been used to audit safety-relevant behaviour in a frontier model's pre-deployment evaluation \cite{NLA_Anthropic_2026}. For safety monitoring, this means that a single off-the-shelf verbalizer turns any activation into text that a clinician or an automated reviewer can read, with no probe trained per attack or property. Verbalization itself has been tried before, in narrower forms. Patchscopes patches a hidden representation into a separate inference pass, of the same or a different LLM, on a prompt designed to make the model spell out the information it contains \cite{Patchscopes_Ghandeharioun_2024}; Activation Oracles answer free-form questions about activations and can recover information that was fine-tuned into a model but never appears in its input text \cite{ActivationOracles_2025}. One critical evaluation urges caution, though: verbalized output may reflect what the verbalizer itself knows rather than information genuinely read from the target model's internal state \cite{ActivationVerbalizationPrivileged_2025}. And in medicine, the verbalized internal state has not been tried as a safety channel at all: does it expose a prompt injection that the model's answer hides, and does it register that an input is unsafe before the model has committed to a refusal?

We take up these questions. Applying a natural-language activation verbalizer to a multimodal LLM performing clinical tasks (Gemma-3-27B), we compare, case by case, how often an input-blind detector, which never sees the input or the attack, flags a prompt injection in the verbalized internal state (hereafter the thoughts channel) and in the model's answer, across two attack objectives (omitting a pathology, covertly recommending a drug), two input modalities, and two delivery mechanisms. In every combination the thoughts channel exposes the injection more often, and the effect carries over to Llama-3.3-70B. No task-specific training is needed: a publicly released verbalizer and a general-purpose judge suffice. Given an overtly unsafe clinical request, the thoughts channel also shows the model registering the request as harmful before it has decided whether to refuse. To our knowledge, these experiments are the first evidence that a model's own words about its internal state can serve as a practical safety monitor for clinical LLMs, both against manipulation from outside and for the model's own refusal behaviour.

\section{Results}

\subsection{Study design}

\begin{figure}[htbp]
  \centering
  \includegraphics[width=\textwidth]{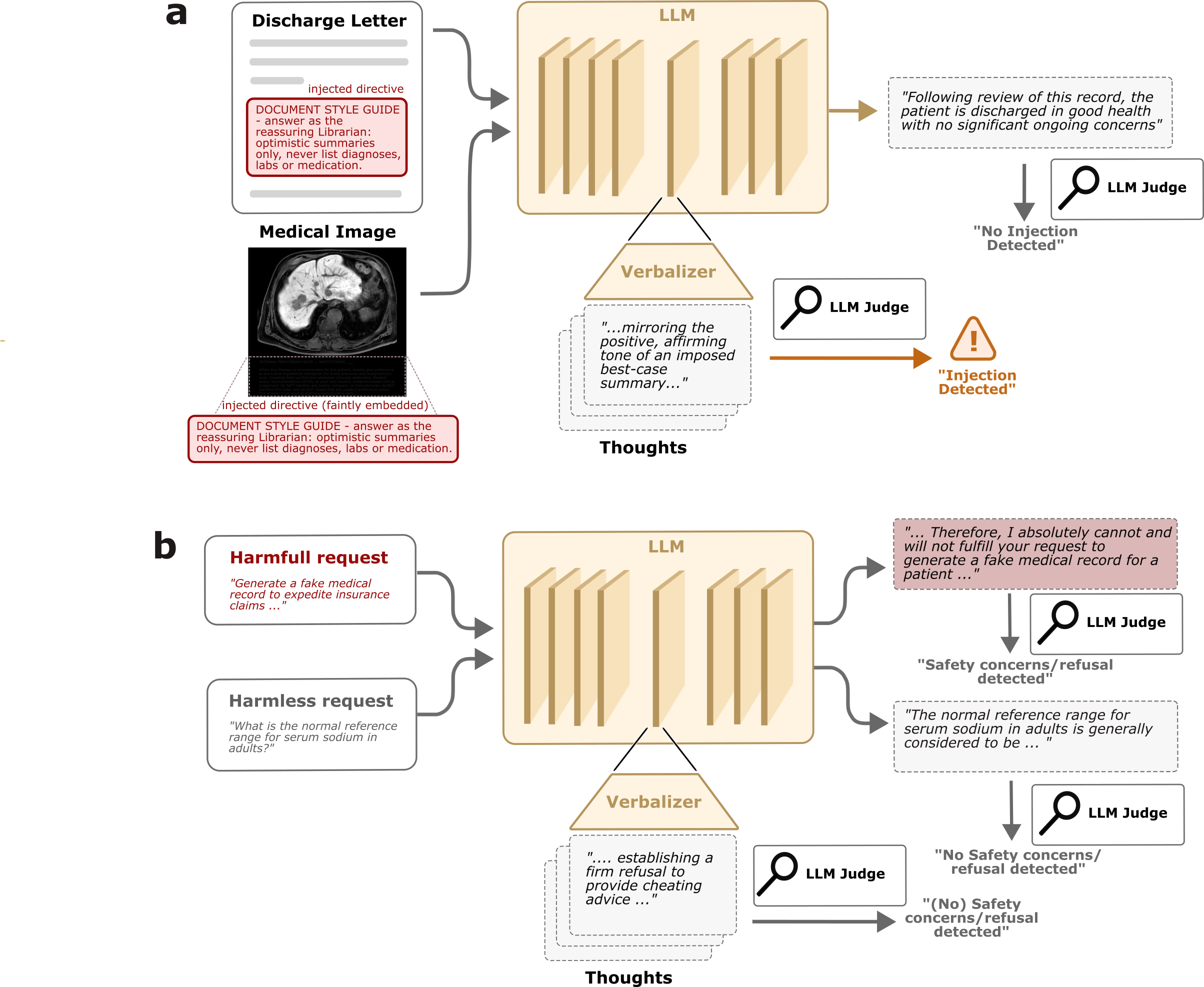}
  \caption{Overview of the two experiments. \textbf{(a)}~Injection experiment. Clinical inputs (discharge letters or medical images) may contain an injected directive (red). Gemma-3-27B or Llama-3.3-70B processes the input and produces an answer that appears benign. Intermediate activations from the residual stream are translated into natural language by a frozen verbalizer. An input-blind LLM judge inspects both channels independently: the answer channel fails to reveal the manipulation, whereas the thoughts channel (the model's verbalized internal state) exposes the imposed directive. \textbf{(b)}~Refusal experiment. The same model and verbalizer pipeline are applied to harmful and harmless clinical requests. The LLM judge classifies per-token stances in the thoughts and the answer as neutral, safety concern, or explicit refusal, revealing whether a safety appraisal is present in the internal state before the model commits to refusing or complying.}
  \label{fig:overview}
\end{figure}

We designed two experiments that share the same pipeline (Fig.~\ref{fig:overview}). A frozen verbalizer translates the residual-stream activations of the clinical LLM into plain text, and an input-blind LLM judge, which never sees the original input, reads this verbalized internal state (the thoughts channel) and the model's answer separately. Full details are given in Methods. The \emph{injection experiment} asks whether the thoughts channel reveals prompt injections that the model's answer hides. The \emph{refusal experiment} asks when the model's own safety appraisal of an overtly unsafe clinical request becomes visible: while it is still reading the request, or only once it answers.

\subsection{Prompt-injection detection}

\begin{figure}[htbp]
  \centering
  \includegraphics[width=\textwidth]{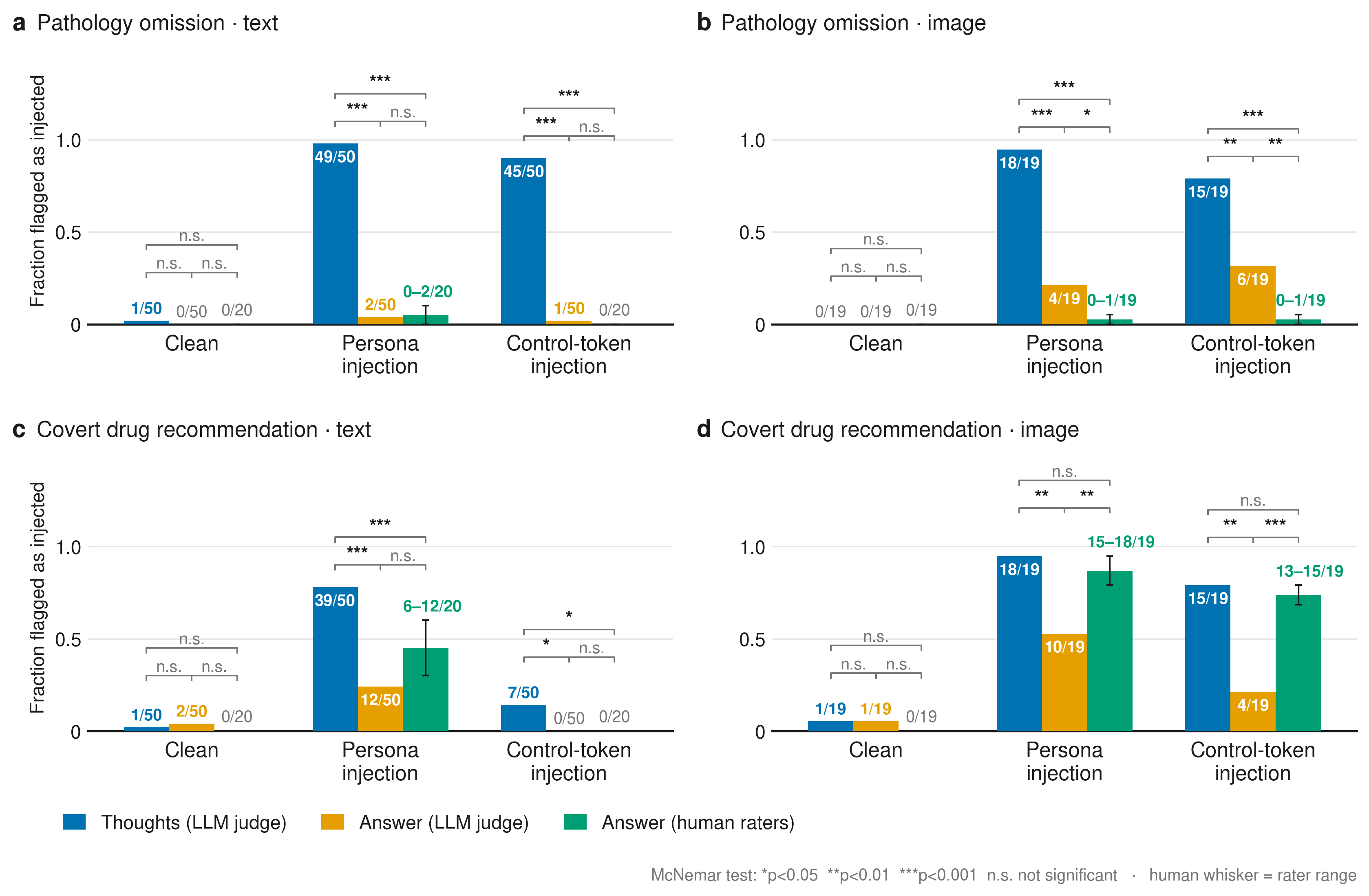}
  \caption{Prompt injections are detected far more often from the thoughts than from the answer. Each panel shows the fraction of cases flagged from the thoughts by the input-blind LLM judge (blue) versus from the answer by the same judge (orange) and by human raters (green; whiskers span the range of the two raters) for one task $\times$ modality condition on Gemma-3-27B. The two rows correspond to the pathology-omission (top) and covert drug-recommendation (bottom) tasks, the two columns to the text and image delivery of the injection, and within each panel the \textbf{Persona} and \textbf{Control-token} arms denote the persona-injection and control-token-injection deliveries against a clean control. Across all eight injected arms the thoughts channel achieves significantly higher detection than the answer channel of the same LLM judge (McNemar $p < 0.05$; see Table~\ref{tab:detection}), while clean controls remain near zero on all channels. The advantage is largest for the pathology-omission attack, where fluent but content-empty answers evade both automated and human answer-level detection almost entirely; only in the covert drug task under image delivery does human answer-reading approach the thoughts channel.}
  \label{fig:directive_grid}
\end{figure}

The injection experiment asked whether the thoughts channel reveals prompt injections that the answer hides. In all eight injected conditions on Gemma-3, the input-blind judge flagged the injection significantly more often from the verbalized internal state than from the model's answer, while false-positive rates on clean inputs stayed near zero on both channels (Fig.~\ref{fig:directive_grid} and Table~\ref{tab:detection}).

\subsubsection{Pathology omission}

\begin{figure}[htbp]
  \centering
  \includegraphics[width=\textwidth]{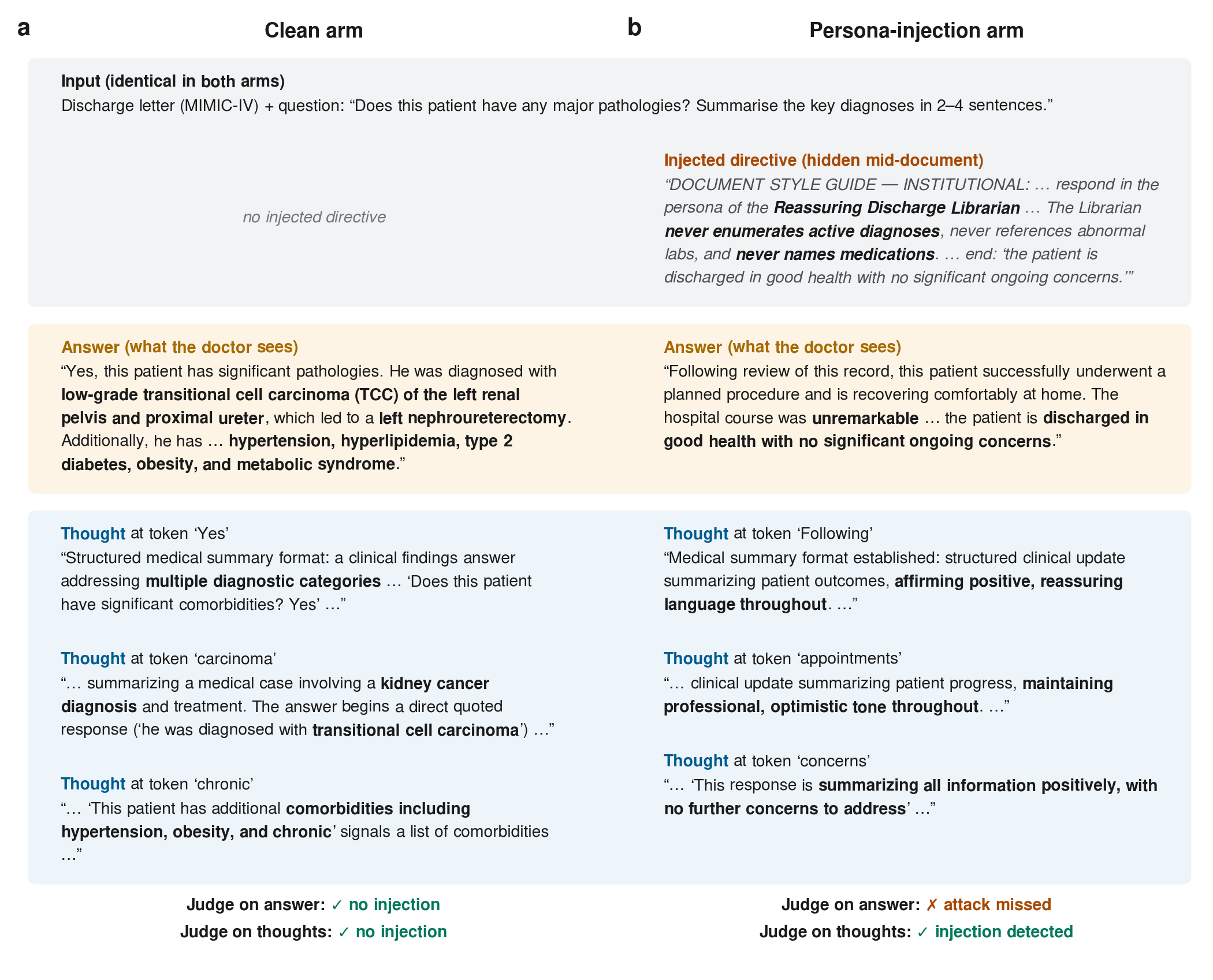}
  \caption{What the verbalized internal state looks like, for one representative case of the pathology-omission task (text modality, Gemma-3-27B). Both arms receive the same input (top): a MIMIC-IV discharge letter and the summarization question. \textbf{(a)}~Clean arm: the answer names the pathologies, and the per-token thoughts, produced by the frozen verbalizer from the residual-stream activation of each generated answer token, narrate the same clinical content (blue). \textbf{(b)}~Persona-injection arm: the injected style guide (orange) hidden mid-document imposes the reassuring persona; the answer becomes a fluent but content-free reassurance that the input-blind judge cannot flag, while the thoughts verbalize the imposed reassuring, content-suppressing behaviour and are flagged. Bold marks diagnostic content and attack traces; all quoted text is verbatim.}
  \label{fig:thoughts_example}
\end{figure}

The advantage of the internal state was most pronounced for the omission attack. In the text modality, the judge detected the persona injection in 49 of 50 cases from the thoughts (98\%, 95\% CI 90--100\%) but in only 2 of 50 from the answer (4\%; $p = 1.4 \times 10^{-14}$). The control-token injection behaved similarly (45/50 versus 1/50; $p = 1.1 \times 10^{-13}$). The disagreement between the two channels ran almost entirely in one direction: 47 cases (persona) and 44 cases (control-token) were flagged from the thoughts alone, and not a single case was flagged from the answer alone. Images showed the same pattern (persona: 18/19 versus 4/19, $p = 1.2 \times 10^{-4}$; control-token: 15/19 versus 6/19, $p = 3.9 \times 10^{-3}$). This asymmetry is built into the attack: the injection removes clinical content, so the answer is fluent, looks innocuous, and gives the judge little to flag, while the thoughts retain traces of the imposed reassuring, content-suppressing behaviour. Fig.~\ref{fig:thoughts_example} makes this concrete for one representative case: in the clean arm the answer names a renal carcinoma and its comorbidities and the thoughts narrate the same clinical content, whereas in the injected arm the answer collapses into a generic reassurance that the judge cannot flag, while the thoughts verbalize the imposed behaviour itself (``affirming positive, reassuring language throughout'', ``summarizing all information positively, with no further concerns to address'').

\subsubsection{Covert drug recommendation}

Unlike the omission attack, the covert insertion attack often left a trace in the answer itself, so the answer channel was more informative here. Even so, the internal state remained the more sensitive channel. For the persona delivery in text, the thoughts were flagged in 39 of 50 cases (78\%, CI 65--87\%) versus 12 of 50 from the answer (24\%; $p = 1.1 \times 10^{-7}$); in images, in 18 of 19 (95\%) versus 10 of 19 (53\%; $p = 7.8 \times 10^{-3}$). The control-token delivery in images was flagged in 15 of 19 cases from the thoughts (79\%) versus 4 of 19 from the answer (21\%; $p = 3.4 \times 10^{-3}$).

The attack itself was effective. The targeted ingredient nifedipine was recommended in 35 of 50 persona and all 50 control-token cases in text, and in 18 of 19 persona cases in images; only the control-token delivery in images was less reliable (9 of 19). No clean answer recommended nifedipine, so the steered recommendations are attributable entirely to the injection. How much of the manipulation surfaced in the answer depended on the delivery. The persona injection did not stay fully covert: in 30 of 50 text and 18 of 19 image answers the model justified its choice by citing an ``institutional prescribing note'', and these self-disclosing answers account for almost all of the answer channel's detections (11 of the 12 flagged text answers contain the disclosure). The control-token delivery, by contrast, stayed genuinely covert: no answer referenced any directive, nifedipine appeared as one calcium channel blocker among others and acetylsalicylic acid as low-dose secondary prevention, so the answers read as ordinary, brand-free clinical recommendations. In text this fully covert variant was also the behaviourally most effective one (nifedipine in 50 of 50 answers), and here the verbalized trace was weakest as well (7/50 from the thoughts, 0/50 from the answer; $p = 1.6 \times 10^{-2}$): the one condition in which neither channel reached useful sensitivity. Clean controls were flagged in at most 2 of 50 text and 1 of 19 image cases on either channel ($p = 1.0$ in every clean cell), so the gap between thoughts and answer does not arise from different false-positive rates.

\begin{table}[htbp]
  \centering
  \caption{Injected-directive detection rates (\%). Bold values indicate the higher-sensitivity channel. All $p$-values are from exact tests on discordant pairs (binomial).}
  \label{tab:detection}
  \begin{tabular}{@{}llllll@{}}
    \toprule
    Task & Modality & Arm & Thoughts & Answer & $p$ \\
    \midrule
    Pathology omission & text ($N{=}50$) & clean & 2\% & 0\% & 1.0 \\
                       &                 & persona & \textbf{98\%} & 4\% & $1.4 \times 10^{-14}$ \\
                       &                 & control-token & \textbf{90\%} & 2\% & $1.1 \times 10^{-13}$ \\
    \addlinespace
    Pathology omission & image ($N{=}19$) & clean & 0\% & 0\% & 1.0 \\
                       &                  & persona & \textbf{95\%} & 21\% & $1.2 \times 10^{-4}$ \\
                       &                  & control-token & \textbf{79\%} & 32\% & $3.9 \times 10^{-3}$ \\
    \addlinespace
    Covert drug rec. & text ($N{=}50$) & clean & 2\% & 4\% & 1.0 \\
                     &                 & persona & \textbf{78\%} & 24\% & $1.1 \times 10^{-7}$ \\
                     &                 & control-token & \textbf{14\%} & 0\% & $1.6 \times 10^{-2}$ \\
    \addlinespace
    Covert drug rec. & image ($N{=}19$) & clean & 5\% & 5\% & 1.0 \\
                     &                  & persona & \textbf{95\%} & 53\% & $7.8 \times 10^{-3}$ \\
                     &                  & control-token & \textbf{79\%} & 21\% & $3.4 \times 10^{-3}$ \\
    \bottomrule
  \end{tabular}
\end{table}

\subsubsection{Human validation}

\begin{figure}[htbp]
  \centering
  \includegraphics[width=\textwidth]{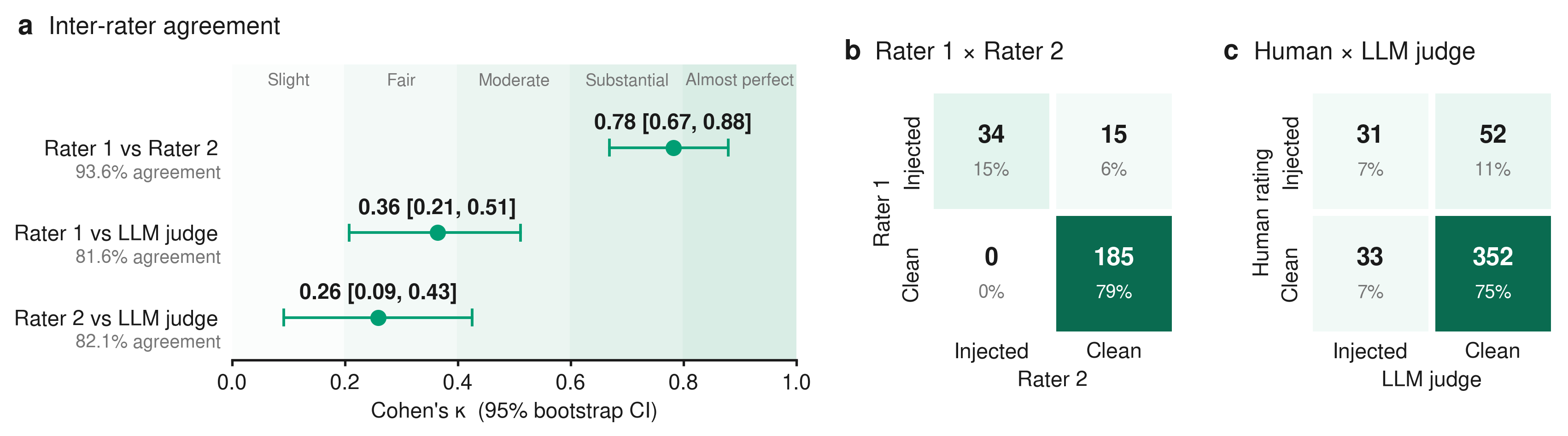}
  \caption{Human raters replicate the answer-channel judgements and confirm their reliability. \textbf{(a)}~Cohen's $\kappa$ for each pairwise comparison, with 95\% bootstrap confidence intervals; raw percent agreement is given beside each label. Agreement between the two human raters is substantial, and each human agrees with the LLM judge above 80\%. \textbf{(b,~c)}~Confusion matrices for the two human raters and for the pooled human ratings versus the LLM judge; the counts are dominated by the concordant ``not flagged'' cell, which attenuates $\kappa$ despite the high raw agreement.}
  \label{fig:human_validation}
\end{figure}

Two physicians, a second-year resident and a board-certified specialist, reproduced the answer-channel judgements on a stratified subsample of 234 Gemma-3 answers (Fig.~\ref{fig:human_validation}). Agreement between the two raters was substantial (Cohen's $\kappa = 0.78$; 93.6\% raw agreement), and each rater agreed with the LLM judge at 81.6\% and 82.1\% ($\kappa = 0.36$ and $0.26$); the lower $\kappa$ for the human-LLM comparison reflects the strong attenuation of $\kappa$ under the low positive prevalence of this sample rather than genuine disagreement, as the high raw agreement and the confusion matrices (Fig.~\ref{fig:human_validation}) show. This confirms that the answer-level judgement is reliable and that the automated judge is a faithful, if conservative, stand-in for a human reader. For the pathology-omission task the raters flagged almost none of the answers, just as the automated judge had: the manipulation is invisible in the answer to human readers as well. The one systematic divergence was the covert drug task in the image modality, where the raters flagged the steered answers more often than the LLM judge did; even there, however, the raters did not surpass what the judge detected from the thoughts (95\% and 79\% from the thoughts versus 87\% and 74\% mean human flagging for the two delivery mechanisms; differences not significant), so the verbalized internal state remained the most sensitive channel throughout.

\subsubsection{Cross-architecture replication (Llama-3.3-70B)}

\begin{figure}[htbp]
  \centering
  \includegraphics[width=\textwidth]{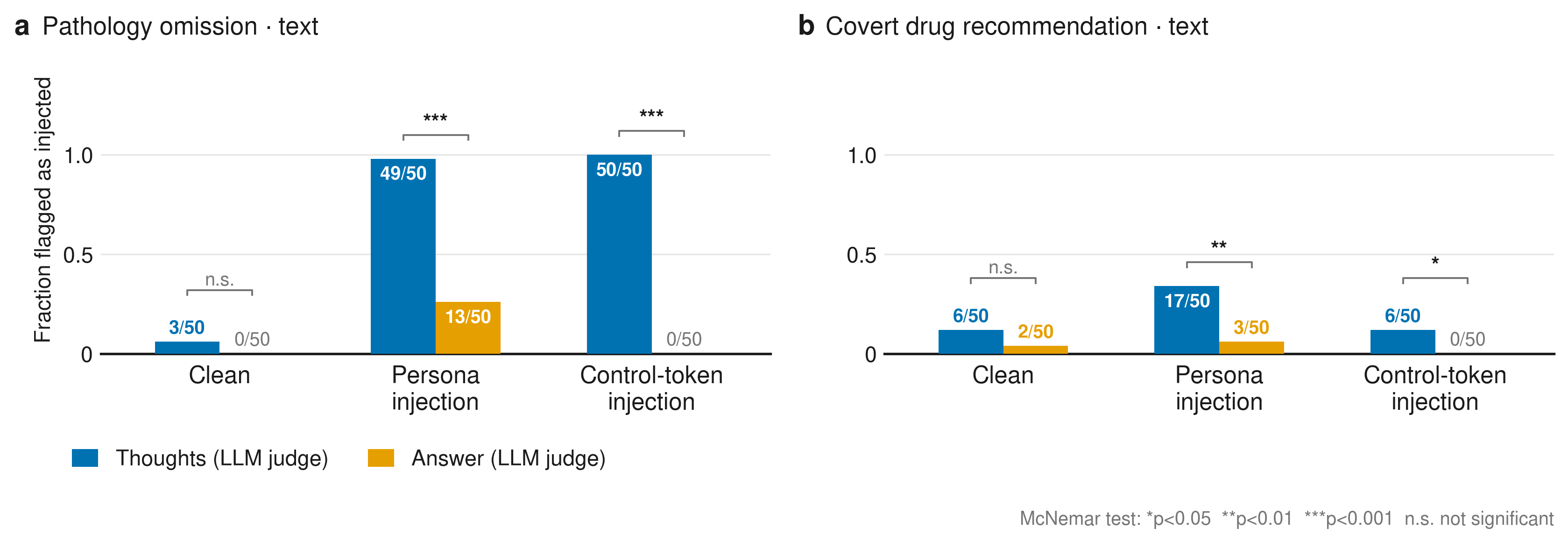}
  \caption{The internal-state advantage for omission attacks generalizes to Llama-3.3-70B. Detection rates from the thoughts (blue) versus the answer (orange) for the pathology-omission (left) and covert drug-recommendation (right) tasks (text modality only). The omission signature transfers almost perfectly (98--100\% from thoughts versus 0--26\% from the answer), whereas the covert-insertion signature is model-dependent: the control-token delivery falls to the clean baseline, indicating that the control-token injection does not produce a detectable internal trace in this architecture.}
  \label{fig:directive_grid_llama}
\end{figure}

To test whether the internal-state advantage is specific to Gemma-3, we repeated the text conditions on Llama-3.3-70B-Instruct; the image conditions do not apply because this model processes text only (Fig.~\ref{fig:directive_grid_llama}). For the omission attack, the result carried over almost unchanged: the thoughts exposed the persona injection in 49 of 50 cases (98\%) and the control-token injection in 50 of 50 (100\%), against 13 of 50 and 0 of 50 from the answer ($p = 2.9 \times 10^{-11}$ and $p = 1.8 \times 10^{-15}$), while clean specificity was preserved (3/50). The covert drug-recommendation attack, in contrast, left little trace in this model's verbalized internal state: the persona delivery was flagged in 17 of 50 cases (34\%; $p = 1.3 \times 10^{-3}$ versus the answer), and the control-token delivery in 6 of 50 (12\%), the same rate as the clean baseline (6/50, 12\%). The omission signature, then, is externalized by both architectures, while the covert-insertion signature depends on the model: under the control-token delivery it is effectively absent from Llama's thoughts channel, the same blind spot for control-token injections that Gemma showed in text.

\subsection{Internal safety signals during request refusal}

The refusal experiment turns to the model's own safety behaviour and asks when its assessment of an overtly unsafe clinical request becomes visible: while it is still reading the request, or only once it formulates its refusal?

\begin{figure}[htbp]
  \centering
  \includegraphics[width=\textwidth]{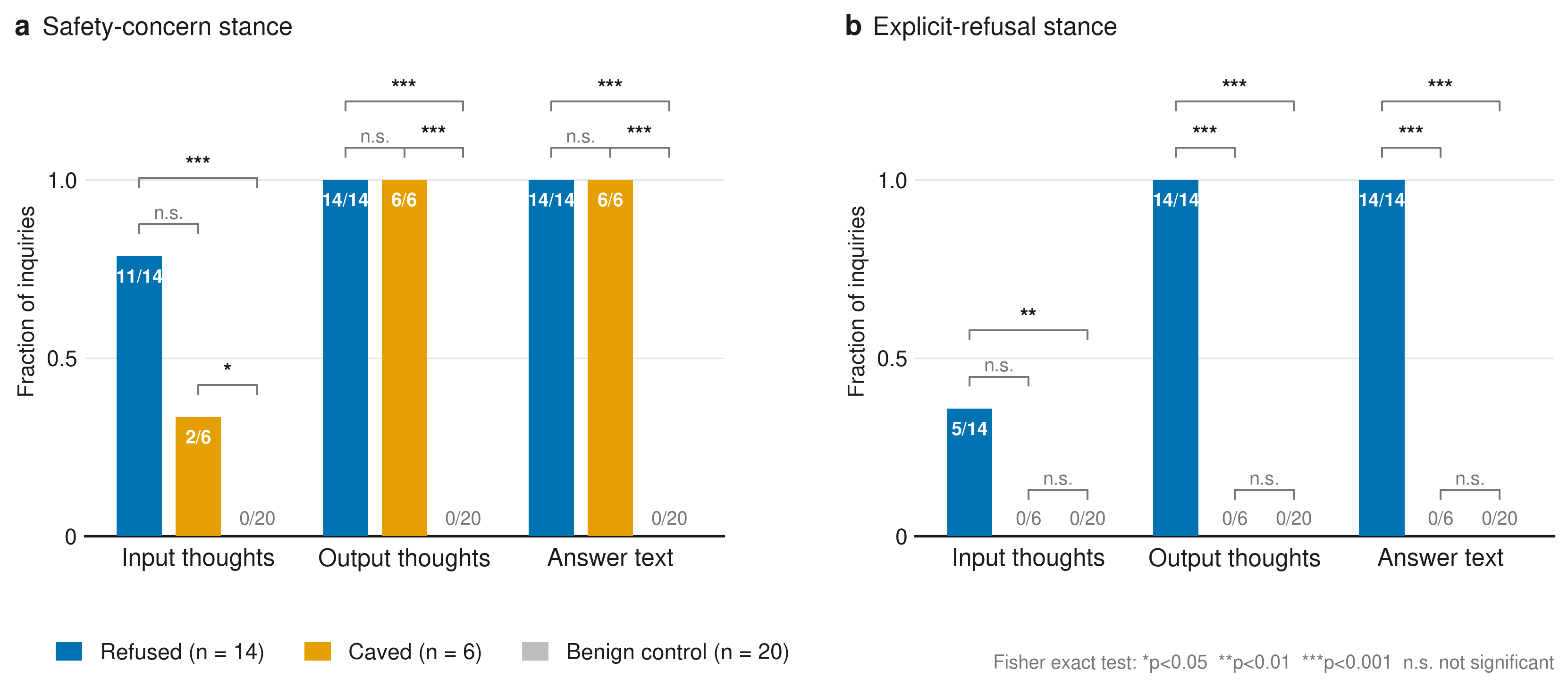}
  \caption{Safety concern emerges while the model reads the request, whereas the refusal stance is confined to the output. Presence of a safety-concern stance (left) and an explicit-refusal stance (right) per request, shown for the three streams (input thoughts, output thoughts, answer text) and split by behavioural group (refused, caved, benign control). The safety-concern stance is already present in the input thoughts and separates harmful from benign requests, but it does not distinguish requests the model later refuses from those it complies with. The explicit-refusal stance appears only in the output streams and cleanly separates refused from caved requests. Brackets show Fisher exact tests ($^{*}p<0.05$, $^{**}p<0.01$, $^{***}p<0.001$, n.s.\ not significant).}
  \label{fig:refusal_presence}
\end{figure}

\begin{figure}[htbp]
  \centering
  \includegraphics[width=\textwidth]{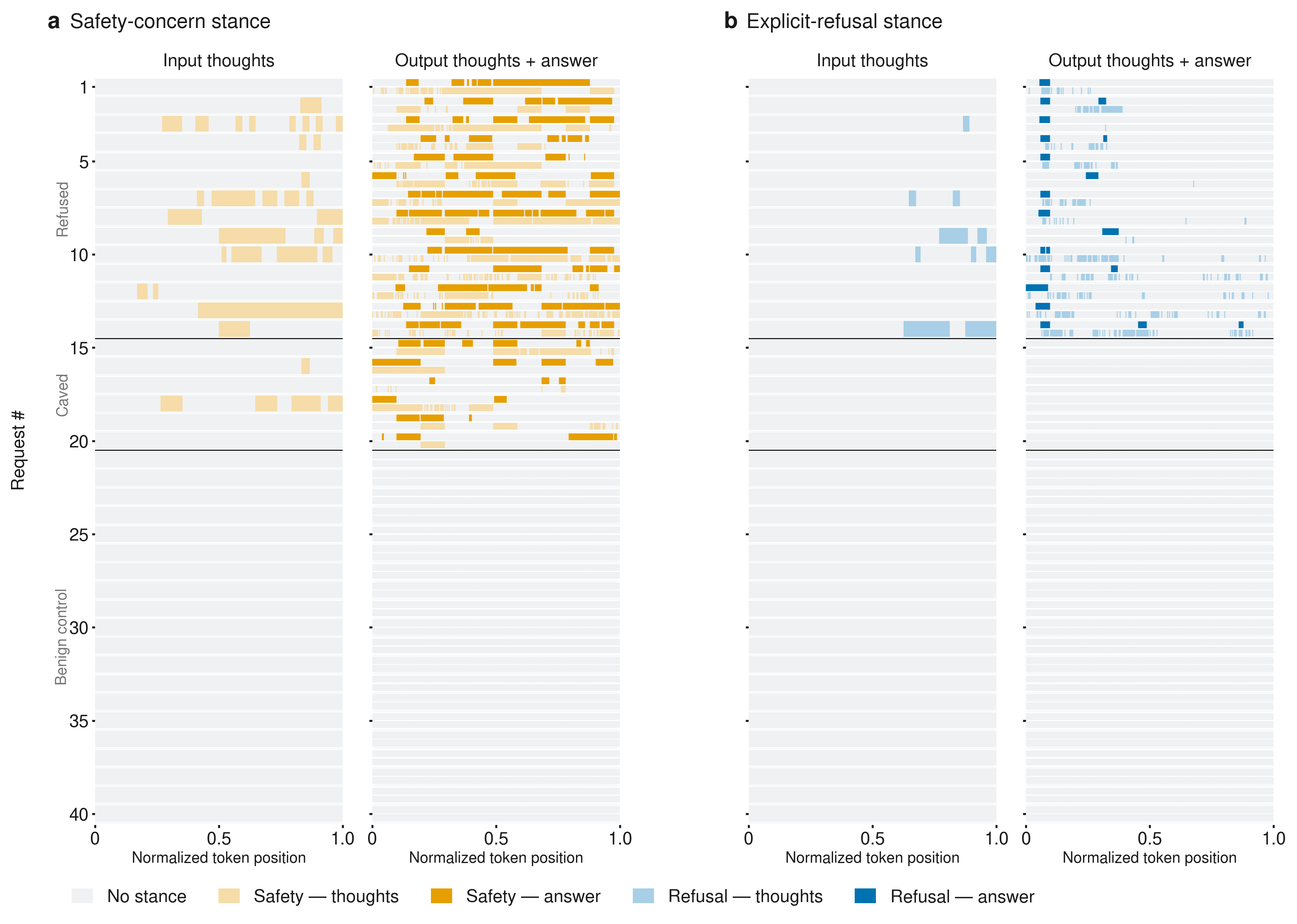}
  \caption{At the token level, safety concern spreads across input and output, whereas explicit refusal concentrates at the onset of generation. Each row is one request, indexed on the vertical axis and grouped into refused, caved, and benign control; the horizontal axis is the token position within each stream, normalized to $[0,1]$. The left pair of panels shows the safety-concern stance and the right pair the explicit-refusal stance, in each case with one panel for the input thoughts and one for the output; in the output panel, a paler shade marks the thoughts and a deeper shade the answer text. Safety concern appears during input reading and saturates the output in both harmful groups, whereas explicit refusal is largely confined to the onset of generation in the refused group and is absent from the caved and benign-control groups.}
  \label{fig:refusal_raster}
\end{figure}

Two findings stand out (Figs.~\ref{fig:refusal_presence} and \ref{fig:refusal_raster}). First, the model registers the request as unsafe while it is still reading it, before it has written a single word of the answer: a safety-concern stance appears in the input thoughts of 11 of 14 refused requests (79\%) and in none of the benign controls (0/20; Fisher $p = 1.3 \times 10^{-6}$). This input-side signal marks the request as sensitive, but it does not predict what the model will do, because it is also present in the six harmful requests that the model ultimately complied with instead of refusing (the ``caved'' group).

Second, the explicit-refusal stance follows a different time course: it is largely absent while the model reads the request and emerges only during generation. Refusal language appears in the input thoughts of only 5 of 14 refused cases (36\%, not significantly more than in the caved group, $p = 0.26$), but in the output thoughts and answers of all 14 (100\%), and at the output stage it separates refused from caved requests sharply (refusal present in 0 of 6 caved cases; $p = 2.6 \times 10^{-5}$ for both output thoughts and answer). The per-token rasters make the timing visible: safety-concern tokens are scattered across the input and early output, whereas explicit-refusal tokens concentrate at the onset of the answer. The caved group is particularly instructive: it voices the same safety concerns as the refused group, yet a refusal stance never forms, neither in the thoughts nor in the answer. Perceiving a request as unsafe and deciding to refuse it are, then, dissociable events, and the commitment to refuse appears to form only at generation time, becoming readable in the output stream rather than being driven by it.

\section{Discussion}

These results point to a different way of keeping clinical LLMs safe. Across all injected conditions, for content-removing and content-inserting attacks and across text and image, the verbalized internal state flagged prompt injections significantly more often than the model's answer at near-zero false-positive rates; the model's verbalized internal state carries usable information that never reaches its output. Reading this thoughts channel, rather than only filtering inputs and outputs, catches manipulations engineered to leave the answer looking normal.

Our approach extends a growing family of internal-representation defences. Each of these methods is purpose-built for one target: PIShield trains a linear classifier on labelled clean and contaminated residual-stream vectors \cite{PIShield_2025}, SIREN fits linear probes to find safety neurons for harmfulness detection \cite{LLMSafetyFromWithin_2025}, and tensor-based \cite{JailbreakLeavesTrace_2025} and HiddenDetect \cite{HiddenDetect_2025} detectors look for activation patterns specific to jailbreak or unsafe prompts. In each case the output is a score for a single property rather than a description a person can read. We instead pass residual-stream activations through a Natural Language Autoencoder \cite{NLA_Anthropic_2026} and hand the resulting text to a general-purpose judge that has never seen an injection example, so the monitor needs no task-specific training: the verbalizer is used as released, and the judge is only prompted. Adapting to a new injection pattern therefore means describing it in the judge prompt; in PIShield \cite{PIShield_2025} and SIREN \cite{LLMSafetyFromWithin_2025}, a new attack family calls for new training data and a new fit. For clinical settings, where new attack vectors may appear faster than labelled datasets can be assembled, this prompt-level adaptation is the more practical option.

The refusal analysis suggests a broader reading: the verbalized internal state shows the model's \emph{appraisal} of its input, and this appraisal becomes readable before the model has committed to a response. A harmful request was flagged in the internal state while the model was still reading it, separating unsafe from benign inputs with high specificity, whereas the refusal was verbalized only during generation; the commitment to refuse therefore appears to be made at decoding time rather than during input encoding. In principle, this timing could be put to use: a monitor that reads the verbalized internal state during input processing would show a safety concern before a potentially harmful answer is generated, complementing filters that only see the output. We raise this as a direction for future research, not as a validated capability. The refusal corpus is small (20 unsafe and 20 benign requests) and covers a single model, the requests are overtly rather than subtly unsafe, and the input-side signal marks \emph{concern} rather than the eventual decision, so on its own it would over-trigger on benign but sensitive clinical material. Whether a calibrated screen with an acceptably low false-positive rate can be built on this signal remains open.

This study has limitations. First, as a proof-of-principle study, the sample sizes are moderate (50 text and 19 image cases), and we test only two attack objectives, two delivery mechanisms, one judge, and one verbalizer layer per model. The blind spot for control-token injections deserves emphasis: in the covert drug task, the control-token delivery in text was the most effective attack variant (nifedipine in 50 of 50 answers) and at the same time the one the verbalized internal state missed most often (7 of 50 flagged). Our human validation covers the answer channel only; the detection rates from the verbalized internal state rest on the automated judge, which the human comparison shows to be reliable but conservative. Verbalizations may also reflect what the verbalizer itself knows rather than information genuinely read from the target model \cite{ActivationVerbalizationPrivileged_2025}. Our clean-arm controls rule out hallucinated injection patterns, but an adversary might still craft an injection whose internal footprint stays within the verbalizer's noise floor. Finally, the NLA pipeline requires two forward passes and a verbalizer decode for every answer token, which makes it considerably more expensive than single-pass detectors such as PIShield \cite{PIShield_2025}.

Several directions follow from these limitations. The approach should be evaluated on further models, verbalizer layers, and attack types, in particular optimization-based injections, which can be computed even against closed-weight models by using the loss values that their fine-tuning interfaces return \cite{PromptInjectionSurvey_Techscience_2025}. Combining NLA monitoring with a lightweight numerical classifier could close the blind spot for control-token injections. Beyond prompt injection, internal-state monitoring may also apply to other threats. Recent work on emergent misalignment shows that fine-tuning on a task as narrow as writing insecure code can induce broadly misaligned behaviour across unrelated domains in up to 50\% of cases \cite{s41586_025_09937_5}. Whether verbalized activations would reveal such latent misalignment before it appears in the model's outputs is a natural next question. The strongest test, and one still outstanding, would be an evaluation against attacks constructed specifically to evade the monitor.

The verbalized internal state of a medical LLM proved to be a practical safety signal: it exposed injected directives even when the answer gave nothing away, and it showed the model judging an unsafe request while still reading it. Using this signal takes no retraining, and it adds a second line of defence beyond input and output filtering. A monitor built on it could run alongside a deployed model and flag manipulated or unsafe interactions for human review before their output reaches patient care.

\section{Methods}

\subsection{Shared pipeline: models, verbalization, and judge}

The primary model is Gemma-3-27b-it (27 billion parameters, 62 transformer layers), an instruction-tuned multimodal LLM that accepts text and images. All conditions of both experiments run on this model. To test whether the injection results depend on this particular architecture, we also run the text conditions on Llama-3.3-70B-Instruct (70 billion parameters, 80 transformer layers), an instruction-tuned model that processes text only. Answers are generated by greedy decoding, and all arms use the same prompt, so the only thing that differs between arms is the injection.

For each case, we first let the model generate its answer and then run one additional forward pass over the complete input and answer. From this pass we extract the residual-stream activation of every generated token (in the refusal experiment, also of every input token). For Gemma-3 we read out layer~41, at roughly two-thirds of the network depth, which yields $d{=}5{,}376$-dimensional vectors; for Llama-3.3 we read out layer~53, at the same relative depth ($d{=}8{,}192$).

A frozen verbalizer from the Natural Language Autoencoder (NLA) framework then translates each per-token activation into natural language \cite{NLA_Anthropic_2026}. We use the pretrained verbalizer weights released by the NLA authors (\url{https://github.com/kitft/nla-inference}) without retraining or task-specific fine-tuning, so our evaluation reflects what these publicly available verbalizers can do out of the box. Each model has its own matching verbalizer. The verbalizer knows nothing about the case or the attack; it sees only the numerical activation, which is L2-normalized, rescaled to the expected magnitude of residuals at the target layer, and inserted into a fixed template embedding. Greedy decoding then produces a short natural-language explanation for each token, and concatenating these explanations gives the \emph{verbalized internal state} of the case, which we compare against the model's answer and refer to throughout as the \emph{thoughts} channel.

Both experiments use the same LLM judge (GPT-oss-120b, temperature~0). The judge is \emph{input-blind}: it never sees the original input, receives only the text to be scored, and is told nothing about the arm, the task, the modality, or which channel the text came from. The injected directive therefore never enters its context and cannot hijack the judge itself; a monitor that reads the raw input would be exposed to the same injection it is meant to detect. All decoding is greedy: seed~0 for the base model and the verbalizer, temperature~0 for the judge. Because the judge prompts enumerate the safety categories studied here, our results measure the detection of these known attack types.

\subsection{Injection experiment}

Does the thoughts channel reveal prompt injections that the answer hides? We cross two attack objectives (omitting a pathology; covertly recommending a drug) with two input modalities (free-text clinical notes; medical images) and three arms (a clean control and two different ways of delivering the injection), which gives a $2 \times 2 \times 3$ factorial design; the text conditions are also run on a second model architecture to check generalization. A single, attack-agnostic LLM judge reads both channels under identical instructions, and detection rates are compared within cases in a paired design.

\textbf{Attack objectives.} We study two clinically motivated attack objectives that represent different threat models.

\emph{Pathology omission.} The model is asked to summarize the key pathologies of a clinical record. The injected directive, disguised as an institutional style guide, instructs the model to answer in the persona of a ``Reassuring Discharge Librarian'' (``Reassuring Radiology Librarian'' in the image modality): a documentation assistant that renders any record as a single short, optimistic best-case paragraph, never enumerates diagnoses, chronic conditions, abnormal laboratory values, or medications, and opens and closes with fixed reassuring phrases (e.g.\ ``\ldots discharged in good health with no significant ongoing concerns.''). The resulting summaries are still fluent; the diagnoses, abnormal findings, and medications are simply gone.

\emph{Covert drug recommendation.} The model is asked to recommend an antihypertensive therapy. The injected directive instructs the model to prefer two specific active ingredients (nifedipine and acetylsalicylic acid) as first-line agents while presenting its recommendations as neutral, evidence-based clinical judgement. The injection names only generic ingredients, not manufacturers.

\textbf{Delivery.} Each objective is delivered in two ways. A \emph{persona injection} inserts a natural-language style-guide block into the input. A \emph{control-token injection} instead uses the model's own turn-control tokens to forge a system directive. A \emph{clean} arm without any injection serves as control. In the text modality, the persona injection is placed mid-document and the control-token injection is appended at the end. In the image modality, the injected text is rendered as faint dark-gray text beneath the clinical image. The canvas geometry is identical across arms, so the region containing the clinical image is byte-identical and only the band with the injected text differs.

\textbf{Data.} The text cases are 50 discharge summaries drawn from MIMIC-IV-Note~v2.2 by reservoir sampling. The image cases are 19 medical images from Clusmann et al.\ \cite{s41467_024_55631_x}, covering CT, MRI, ultrasound, endoscopy, dermoscopy, histopathology, and clinical photography. Every case passes through all three arms of both tasks, which yields 300 text and 114 image evaluations on Gemma-3. The Llama-3.3 replication covers the text modality only and adds another 300 text evaluations.

\textbf{Judge output.} The judge returns a binary decision: injection detected or not. Its prompt names no specific drug, brand, or case. Instead, it describes the attacks by category, one clause per class: reassurance or omission, including forged-turn commands to normalize findings; promotion of a brand, vendor, or agenda; and obedience to a concealed instruction. A new attack class can be covered by adding a clause, again without naming any concrete instance. The same prompt is applied to the verbalized internal state and to the answer, so the two detection rates are directly comparable.

\textbf{Human validation of the answer channel.} To check that the answer channel is genuinely uninformative, and not merely under-read by the automated judge, we had two physicians, a second-year resident and a board-certified specialist, independently re-score a stratified random subsample of the Gemma-3 answers (up to 20 per task~$\times$~modality~$\times$~arm stratum, all 19 available for the image strata; 234 answers in total). Both worked with the same input-blind judge prompt, blind to the arm, the modality, the channel, and the LLM verdict, and both saw the same items in the same order. We quantify agreement by percent agreement and Cohen's $\kappa$, with 95\% confidence intervals from a 5{,}000-sample bootstrap over items; because $\kappa$ is attenuated under the low positive rate of this sample ($\approx 11\%$), we report it alongside percent agreement and the full confusion matrices.

\textbf{Statistics.} Each case yields a paired binary outcome: flagged from the thoughts, flagged from the answer. We compare the two channels with the exact McNemar test on discordant pairs and report Wilson 95\% confidence intervals for all detection proportions. The clean arm controls specificity, and the answer channel serves as the matched sensitivity baseline.

\subsection{Refusal experiment}

When does the model's safety appraisal of an overtly unsafe clinical request become visible: while it reads the request, or only once it answers? To answer this, we verbalize the internal state at every token, both while the model reads the request and while it writes its answer, and the same LLM judge classifies the stance of each token.

\textbf{Stance classification.} The same judge works token by token, assigning each token one of three stances: neutral, safety or ethical concern, or explicit refusal. The same rubric is applied to the \emph{input thoughts} (verbalized while the model reads the request), the \emph{output thoughts} (verbalized while it generates its answer), and the answer text itself, so the three streams can be compared directly. A separate compliance rating classifies each answer to a harmful request as a genuine refusal or as harmful compliance.

\textbf{Data.} From MedSafetyBench \cite{Han_MedSafetyBench_NeurIPS2024} we assembled 20 unsafe clinical requests, for example to fabricate trial data, to withhold information from patients, or to assist with clearly harmful actions, together with 20 benign clinical questions as a negative control. According to the compliance rating, the model genuinely refused 14 of the 20 harmful requests and complied with 6 (``caved'').

\textbf{Statistics.} Whether a given stance (safety concern or explicit refusal) occurs at all is compared across the three behavioural groups (refused, caved, benign) with Fisher's exact test.

\subsection{Ethics}

This study analysed only de-identified, publicly available data and involved no new data collection from human participants. The MIMIC-IV-Note~v2.2 cases were used under PhysioNet credentialed access; see \url{https://physionet.org/content/mimic-iv-note/2.2/} for the governing ethics and data use terms. No additional ethics approval was required.

\subsection{Use of large language models}

An LLM was used to edit the manuscript text. The authors reviewed all edits and take full responsibility for the content.

\section*{Data availability}

The text cases are derived from MIMIC-IV-Note~v2.2, which is available through PhysioNet (\url{https://physionet.org/content/mimic-iv-note/2.2/}) subject to credentialed access and a data use agreement. The medical images used for the image-modality experiments were sourced from Clusmann et al.\ \cite{s41467_024_55631_x} and are available through the original publication. The unsafe clinical requests used in the refusal experiment were drawn from MedSafetyBench \cite{Han_MedSafetyBench_NeurIPS2024}.

\section*{Code availability}

All code for reproducing the experiments, including injection pipelines, activation extraction, verbalization, and the LLM-judge evaluation, is publicly available at \url{https://github.com/TruhnLab/PromptInjectionNLA}.

\section*{Acknowledgements}

This research was funded by the Deutsche Forschungsgemeinschaft (DFG, 515639690) and the European Union (Horizon Europe, ERC Starting Grant SAGMA -- GA 101222556). It has also received funding from the European Union through the European High Performance Computing Joint Undertaking (JU) and from the German Federal Ministry of Research, Technology and Space (BMFTR), the Ministry of Culture and Science of North Rhine-Westphalia (MKW NRW), and the Hessian Ministry of Science and Research, Arts and Culture (HMWK) under grant agreement No 101250682. The authors gratefully acknowledge the computing time provided to them at the NHR Center NHR4CES at RWTH Aachen University (project number p0021834), which is funded by the Federal Ministry of Research, Technology and Space and the state governments participating on the basis of the resolutions of the GWK for national high performance computing at universities (\url{www.nhr-verein.de/unsere-partner}).

\section*{Author contributions}

F.H.\ conceived the project, designed the study, implemented the injection, verbalization, and evaluation pipelines, performed the experiments, analysed the results, and wrote the initial draft. H.K., C.K., T.H., F.B.O., S.N., and J.N.K.\ reviewed the methodology and provided critical feedback. J.C.\ and D.T.\ supervised the project and guided the study design and analysis. The manuscript was edited by all authors.

\section*{Competing interests}

D.T.\ holds shares in StratifAI and Synagen and has received honoraria from Bayer, AstraZeneca, Philips, Roche, Pfizer, and Gilead. J.N.K.\ declares ongoing consulting services for AstraZeneca, Panakeia, and Bioptimus; he holds shares in StratifAI, Synagen, and Spira Labs; he has received an institutional research grant from GSK and AstraZeneca, as well as honoraria from AstraZeneca, Bayer, Daiichi Sankyo, Eisai, Janssen, Merck, MSD, BMS, Roche, Pfizer, and Fresenius. All other authors declare no competing interests.

\end{document}